\pdfoutput=1

\documentclass[letterpaper, 10 pt, conference]{ieeeconf}  

\usepackage{graphicx}
\usepackage{comment}
\usepackage{amsmath,amssymb} 
\usepackage{color}
\usepackage{url}

\usepackage{algorithm}
\usepackage{algpseudocode}

\usepackage{soul}
\usepackage{xspace}
\makeatletter
\DeclareRobustCommand\onedot{\futurelet\@let@token\@onedot}
\def\@onedot{\ifx\@let@token.\else.\null\fi\xspace}
\def\eg{\emph{e.g}\onedot} 
\def\ie{\emph{i.e}\onedot} 
 
\def\etc{\emph{etc}\onedot} 
 
\def\etal{\emph{et al}\onedot}

\usepackage{amsmath,xparse} 
\usepackage{xcolor}

\newcommand{\Tref}[1]{Table~\textcolor{blue}{\ref{#1}}}

\newcommand{\Fref}[1]{Fig.~\textcolor{blue}{\ref{#1}}}
\newcommand{\Sref}[1]{Sec.~\textcolor{blue}{\ref{#1}}}

\usepackage{epsfig}
\usepackage{graphicx}

\usepackage[symbol]{footmisc}

\usepackage{multirow}

\newcommand{\RNum}[1]{\uppercase\expandafter{\romannumeral #1\relax}}

\IEEEoverridecommandlockouts                              

\begin{document}

\title{\LARGE \bf
Volumetric Propagation Network: \\ Stereo-LiDAR Fusion for Long-Range Depth Estimation
}

\author{Jaesung Choe$^{1}$, Kyungdon Joo$^{2}$, Tooba Imtiaz$^{3}$ and In So Kweon$^{3}$
\thanks{This work (K. Joo) was supported by Institute of Information \& communications Technology Planning \& Evaluation (IITP) grant funded by the Korea government (MSIT) (No.2020-0-01336, Artificial Intelligence Graduate School Program (UNIST)). \textit{(Corresponding author: I. S. Kweon.)}
}
\thanks{$^{1}$J. Choe is with the Division of the Future Vehicle, KAIST, Daejeon 34141, Republic of Korea.
{\tt jaesung.choe@kaist.ac.kr}}
\thanks{$^{2}$K. Joo is with the Artificial Intelligence Graduate School and the Department of Computer Science, UNIST, Ulsan 44919, Republic of Korea.
{\tt kdjoo369@gmail.com, kyungdon@unist.ac.kr}}
\thanks{$^{3}$T. Imtiaz, and I. S. Kweon are with the School of Electrical Engineering, KAIST, Daejeon 34141, Republic of Korea.
{\tt \{timtiaz, iskweon77\}@kaist.ac.kr}}
\thanks{This paper is for the presentation in ICRA 2021.}
}

\markboth{IEEE ROBOTICS AND AUTOMATION LETTERS. PREPRINT VERSION. ACCEPTED March, 2021}
{Choe \MakeLowercase{\textit{et al.}}: Volumetric Propagation Network: Stereo-LiDAR Fusion for Long-Range Depth Estimation}

\maketitle

\begin{abstract}
Stereo-LiDAR fusion is a promising task in that we can utilize two different types of 3D perceptions for practical usage -- dense 3D information~(stereo cameras) and highly-accurate sparse point clouds~(LiDAR). However, due to their different modalities and structures, the method of aligning sensor data is the key for successful sensor fusion.
To this end, we propose a geometry-aware stereo-LiDAR fusion network for long-range depth estimation, called \emph{volumetric propagation network}. The key idea of our network is to exploit sparse and accurate point clouds as a cue for guiding correspondences of stereo images in a unified 3D volume space. Unlike existing fusion strategies, we directly embed point clouds into the volume, which enables us to propagate valid information into nearby voxels in the volume, and to reduce the uncertainty of correspondences. Thus, it allows us to fuse two different input modalities seamlessly and regress a long-range depth map. Our fusion is further enhanced by a newly proposed feature extraction layer for point clouds guided by images: \emph{FusionConv}. FusionConv extracts point cloud features that consider both semantic~(2D image domain) and geometric~(3D domain) relations and aid fusion at the volume. Our network achieves state-of-the-art performance on the KITTI and the Virtual-KITTI datasets among recent stereo-LiDAR fusion methods. 
\end{abstract}

\section{Introduction}
\label{sec:Introduction}
Sensor fusion is the process of merging data from multiple sensors, which makes it possible enrich the understanding of 3D environments (\ie, 3D perception) for autonomous driving or robot perception. Each sensor has its unique properties and can complement other sensors' limitations by fusion. In particular, sensor fusion -- such as two cameras (stereo matching) or LiDAR and a single camera (depth completion) -- allows us to estimate accurate depth information. Several studies have explored fusion-based depth estimation algorithms~\cite{geometry,semi_global_matching,displets}. The quality of depth estimated by these traditional methods has been further improved with the advent of deep learning~\cite{dispnet,gcnet,sparse-to-dense,guidenet,stereo_object,seg2reg}. 

Recently, stereo-LiDAR fusion has been getting more attention for practical usage in autonomous driving~\cite{stereolidar_norm_costV_ccvn,stereolidar_00,stereolidar_01}. Compared to traditional fusion-based depth estimation, stereo-LiDAR fusion is a novel task where we can utilize two different types of 3D perception: dense 3D information from stereo cameras and sparse 3D point clouds from LiDAR, which can further improve the depth quality. Within this fusion framework, aligning different sensor data into a unified space is an essential step to fully operate depth estimation task. Previous works~\cite{stereolidar_norm_costV_ccvn,stereolidar_00,stereolidar_01} perform fusion on the 2D image domain using the geometric relationship between sensors (\ie,~intrinsic and extrinsic parameters). For example, Wang~\etal~\cite{stereolidar_norm_costV_ccvn} project point clouds into the 2D image domain to align images with sparse depth maps of the projected point clouds. 
%
%
However, because neighboring pixels in a 2D image space are not necessarily adjacent in the 3D space, this 2D fusion in the image domain may lose depth-wise spatial connectivity, thus have difficulty in estimating accurate depth at distant regions.
%
%
For geometry-aware fusion of stereo images and point clouds, it is necessary to maintain the spatial connectivity in a unified 3D space.

\begin{figure}[!t]
    \centering
    \includegraphics[width=1.0\linewidth]{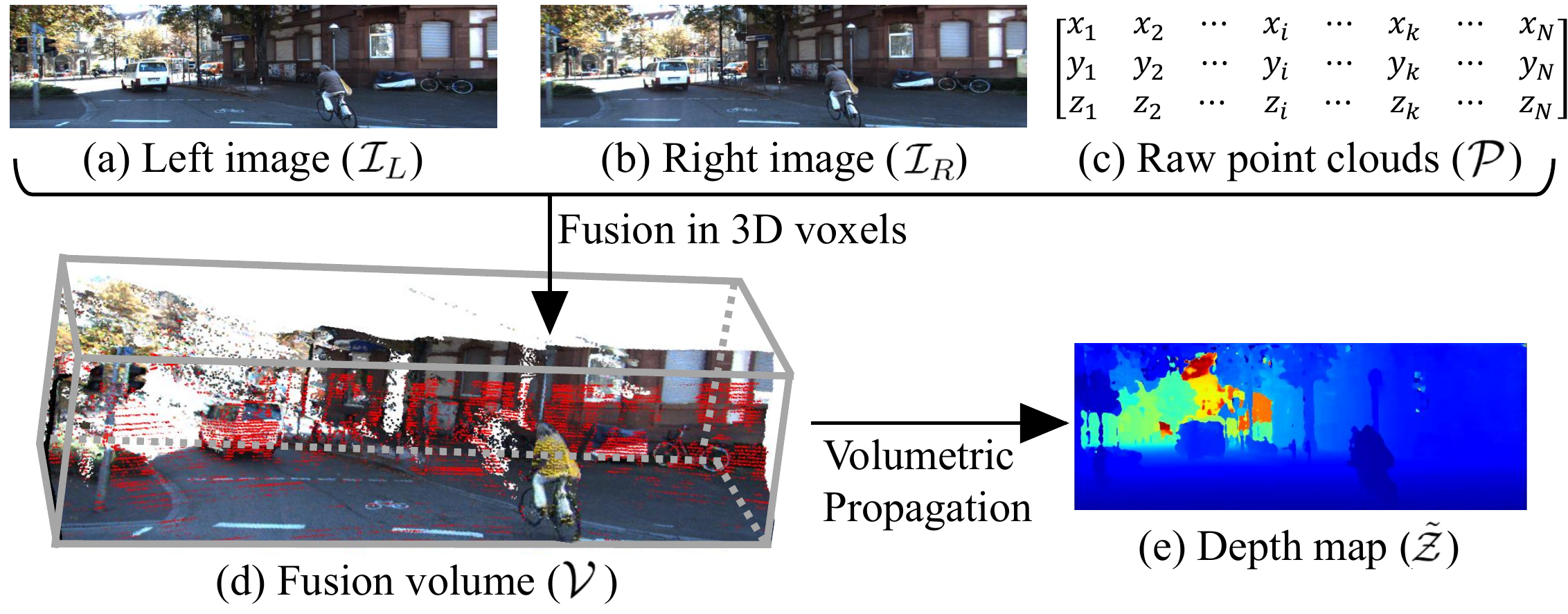}
    \vspace{-7mm}
    \caption{\textbf{Pipeline of the proposed stereo-LiDAR fusion.} 
    Given (a,b)~stereo images and (c)~point clouds, we fuse two different input modalities at 3D voxels of (d) fusion volume, which is a unified 3D volumetric space, to infer (e)~depth map by volumetric propagation. }
    \vspace{-4mm}
    \label{fig:fig_intro}
\end{figure}


In this paper, we propose a geometry-aware stereo-LiDAR fusion network for long-range depth estimation, called \emph{volumetric propagation network}. To this end, we define 3D volumetric features, called \emph{fusion volume}, as a~unified 3D volumetric space for fusion, where the proposed network computes the correspondences from both stereo images and point clouds, as shown in~\Fref{fig:fig_intro}. Specifically, sparse points become seeds of correct correspondence to reduce the uncertainty of matching between the stereo images. Then, our network propagates this valid matching to the overall volume and computes the matching cost of the rest of the volume through stereo images. 
Furthermore, we facilitate the volumetric propagation by embedding point features into the fusion volume. The point features are extracted by our image-guided feature extraction layer from raw point clouds, called \emph{FusionConv}. Our FusionConv considers both semantic (2D image domain) and geometric (3D domain) relation for the tight fusion of image features and point features. Finally, our approach achieves state-of-the-art performance among stereo-LiDAR fusion methods for the KITTI dataset~\cite{kitti-completion} and the Virtual-KITTI dataset~\cite{virtual-kitti}.


\section{Related work}
\label{sec:Related works}
We review sensor fusion-based depth estimation methods according to types of sensor systems: stereo cameras, mono-LiDAR fusion and stereo-LiDAR fusion.

\vspace{1mm} \noindent \textbf{Stereo cameras.} \ 
Stereo matching is a task that reconstructs the 3D environment captured from a pair of cameras~\cite{geometry}. By computing the dense pixel correspondence between a rectified image pair, stereo matching infers the inverse depth \ie, disparity map. Recently, deep-learning techniques have paved the way for more accurate and robust matching by using volume-based deep architectures~\cite{gcnet,psmnet,dpsnet}. 
Cost volume is one of the popular volumetric representations that encompass the 3D space in the referential camera view\footnote{We set the left camera as the referential~\cite{geometry}.} along the disparity axis. This property is beneficial for computing matching cost between stereo images. 
Despite the large improvement, stereo matching still lacks accurate depth estimation at distant regions. 
To address this issue, LiDAR is a prominent sensor to estimate long-range depth.

\vspace{1mm} \noindent \textbf{Mono-LiDAR fusion.} \ 
Depth completion is a task that estimates a depth map using a monocular camera and a LiDAR sensor. By propagating highly accurate but sparse points from a LiDAR sensor, this task aims to densely complete the depth map with the help of image information. Recent deep-learning-based  methods~\cite{guidenet,sparse-to-dense,2D-3D-fusion-depth-completion,jinsunpark} largely increase the quality of depth. These methods~\cite{guidenet,sparse-to-dense,jinsunpark} employ a pre-processing stage that projects point clouds into image domain, and feed this sparse depth map as input to a network~(\ie,~early fusion). On the other hand, Chen~\etal~\cite{2D-3D-fusion-depth-completion} introduce an intermediate fusion scheme in the feature space. They initially extract features from each modality and implement the fusion in the image feature space by projecting point features. Despite the improvement thus achieved, depth completion task has difficulties in estimating depth at unknown areas not encompassed by point clouds.

\vspace{1mm} \noindent \textbf{Stereo-LiDAR fusion.} \ 
Stereo camera-LiDAR fusion (simply denoted as stereo-LiDAR fusion) has been recently demonstrated to further increase the accuracy of depth estimation by utilizing additional sensory information. Relying on the highly accurate but sparse depth map from point clouds,
Park~\etal~\cite{stereolidar_00} refine the disparity map from stereo cameras using a sparse depth map. Despite the increased quality of estimated depth, the fusion within this method lies in 2D image domain and is therefore insufficient to maintain metric accuracy of point clouds for long-range depth estimation. Another recent work, Wang~\etal~\cite{stereolidar_norm_costV_ccvn}, introduce the idea of input fusion (\ie, stereo RGB-D input) and volumetric normalization conditioned by sparse disparity (\ie, projected point clouds). Typically, this conditional cost volume normalization~\cite{stereolidar_norm_costV_ccvn} mainly affects the 3D volumetric aggregation and looks closer to the idea of our volumetric propagation. However, this method has difficulty in estimating the accurate depth in remote area.

To address this issue, we introduce the volumetric propagation network that aims to fuse the two input modalities: stereo images and point clouds in a unified 3D volume space, fusion volume. Within the fusion volume, we regard the sparse points as the seed of valid matching between stereo images and propagate the valid matching to the overall volume space. To do so, our network (1) maintains metric accuracy of point clouds during fusion and (2) reduces the uncertainty of stereo matching where point clouds do not exist. We further facilitate fusion at the fusion volume by our feature extraction layer for point clouds, FusionConv. Among recent stereo-LiDAR fusion works~\cite{stereolidar_01,stereolidar_00,stereolidar_norm_costV_ccvn}, with these two contributions, we achieve remarkable state-of-the-art depth estimation performance.

\section{Overview}
\label{sec:Overview}

\begin{figure*}[!t]
\vspace{+3mm}
\centering
\includegraphics[width=0.85\linewidth]{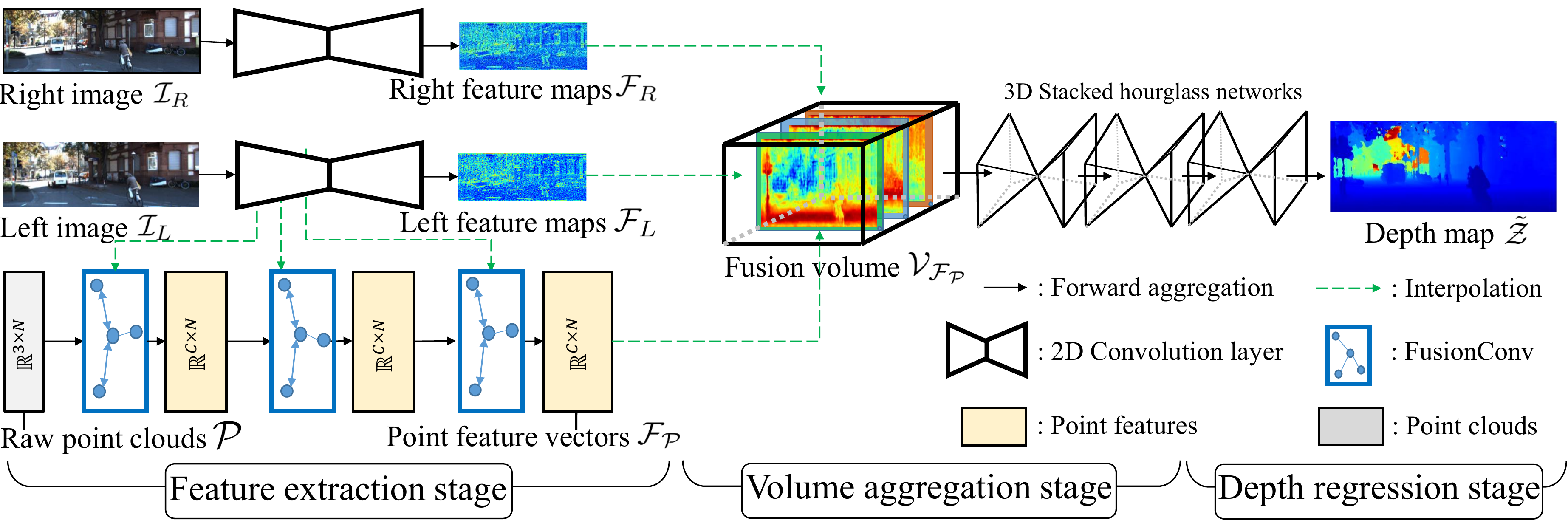}
\caption{\textbf{Overall architecture.} Our network consists of three stages: feature extraction stage, volume aggregation stage, and depth regression stage. We initially extract features from input modalities. In particular, we use FusionConv layers to infer point feature vectors ${\mathcal{F}_{\mathcal{P}}}$. These extracted features are embedded into the fusion volume $\mathcal{V}_{\mathcal{F}_{\mathcal{P}}}$ to compute the correspondence {with stereo feature maps $\mathcal{F}_{L}$, $\mathcal{F}_{R}$}. After volume aggregation through 3D stacked hourglass networks, we finally obtain the depth map $\widetilde{Z}$ in the referential camera view.
}
\label{fig:fig_arch}
\vspace{-4mm}
\end{figure*}

We aim to estimate a long-range, highly-accurate, and dense depth map $\widetilde{\mathcal{Z}}$ in the referential camera viewpoint from rectified stereo images $\mathcal{I}_{L}$ and $\mathcal{I}_{R}$, LiDAR point clouds $\mathcal{P}$ and their calibration parameters (\eg, camera intrinsic matrix $\mathbf{K}$). The point clouds are presented as a set of $N$ number of 3D points $\mathcal{P} {=} \{ \mathbf{p}_{i} \}_{i=1}^{N}$, where each point $\mathbf{p} {=} [x, y, z]^\top$ lies in the referential camera coordinates.\footnote{For simplicity, we transform the point clouds in the LiDAR coordinates into the referential camera coordinates using extrinsic parameters.} Under the known $\mathbf{K}$, we can project each point into the image domain, $\mathbf{x} \simeq \mathbf{K}\mathbf{p}$, where the projected image point $\mathbf{x}$ can be located at the pixel location of $(u, v)$. Using this geometric relation in conjunction with interpolation, we fuse the two modalities at the fusion volume~(\Sref{sec:Volumetric Propagation}) and FusionConv~(\Sref{sec:Fused Convolution Layer}). The overview of the proposed approach is presented in \Fref{fig:fig_arch}. 

\section{Volumetric Propagation Network}
\label{sec:Volumetric Propagation}

In this section, we detail how we construct our fusion volume $\mathcal{V}$ to fuse two different input modalities, stereo images and raw point clouds, in a manner of volumetric propagation. The proposed fusion volume~$\mathcal{V}$ is a unified 3D volumetric space for stereo-LiDAR fusion to estimate long-range depth map $\widetilde{\mathcal{Z}}$ in the referential view. Unlike the traditional voxel representation -- cost volume~\cite{gcnet,psmnet,ganet} which quantizes the 3D data along the disparity axis, 
our fusion volume~$\mathcal{V}$ describes the 3D environment as an evenly distributed grid voxel space along the depth range~(\ie, metric scale). This is to reduce the quantization loss when we embed sparse points $\mathcal{P}$ into the fusion volume. While embedding points into the cost volume dramatically increase the quantization loss in distant regions, our metric-scale fusion volume does not. Specifically, we define the fusion volume~$\mathcal{V}$ as a 3D volume representation and each voxel~$\mathbf{v} \in \mathcal{V}$  contains a feature vector of a certain dimension. That is, $\mathcal{V} \in \mathbb{R}^{W \times H \times D \times (2C+1)}$, where $W$, $H$ and $D$ denote the number of voxels along the width, height, and depth axes, respectively, and $C$ represents the dimension of the image feature vector. Note that the feature dimension of the volume~$\mathcal{V}$ is set as $(2C+1)$ to embed stereo image features and point cloud features along the different channels but in a unified volume $\mathcal{V}$.

In the constructed fusion volume, which is empty at the beginning, we first fill in the stereo information. Inspired by the differential cost volume~\cite{gcnet,psmnet}, filling the volume with features from sensory data enables the end-to-end learnable voxel representation to estimate depth. From stereo images $\mathcal{I}_{L}$, $\mathcal{I}_{R} \in \mathbb{R}^{4W \times 4H \times 3}$, we extract stereo feature maps $\mathcal{F}_{L}$, $\mathcal{F}_{R} \in \mathbb{R}^{W \times H \times C}$ using feature extraction layers from our baseline method~\cite{psmnet}. By projecting the pre-defined location of each voxel~$\mathbf{v}$ into the image domain, we fill the fusion volume~$\mathcal{V}$ with stereo features (see \Fref{fig:fig_arch}). A precise description of the fusion volume composition is included in the supplementary material.

Our fusion volume encapsulates the 3D environments linearly to the metric scale as depth volumes do~\cite{pseudo_lidar,dsgn}. However, while traditional depth volumes~\cite{pseudo_lidar,dsgn} are the results of transformation from the initially-built cost volume (\ie,~from disparity to depth), our fusion volume directly built from sensory features. Since the quantization loss happens when points are embedded into the initially-built volume, direct construction of a metric-scale volume can reduce the quantization loss of points, especially at a farther area. For this purpose, with the known camera matrix~$\mathbf{K}$ and the pre-defined location of each voxel in the volume $\mathcal{V}$, we are able to embed point clouds $\mathcal{P}$ into the volume~$\mathcal{V}$ as binary representation with less quantization loss at farther area. Voxels embedded by points are filled with 1 (occupied) and the others are filled with 0 (non-occupied or empty). To do so, our fusion volume maintains the geometric and spatial relation of stereo images and point clouds within a unified volumetric space.

So far, we have incorporated stereo feature maps~$\mathcal{F}_{L}$, $\mathcal{F}_{R}$ and raw point clouds~$\mathcal{P}$ into the fusion volume $\mathcal{V}$. With this volume~$\mathcal{V}$, we can propagate the embedded points to the overall volume for computing the matching cost between stereo features ($\mathcal{F}_{L} \text{, } \mathcal{F}_{R}$) or among stereo features and point clouds ($\mathcal{F}_{L} \text{, } \mathcal{F}_{R} \text{, } \mathcal{P}$) through the following 3D convolution layers in stacked hourglass networks (see \Fref{fig:fig_arch}). Meanwhile, the 3D convolution layers compute both the spatial correspondence and the channel-wise features within the volume $\mathcal{V}$, so that channel-wise information is also an one important factor in our metric-aware fusion. To further facilitate the fusion, we discuss feature extraction from raw point clouds $\mathcal{P}$ in the next section.

\section{Fused Convolution Layer}
\label{sec:Fused Convolution Layer}

Recently, many research works~\cite{pointnet,continuous_conv,point_conv,dynamic_graph_conv,interp_conv} have proposed feature extraction layers for point clouds and have demonstrated the potential of leveraging local neighbors in aggregating point feature vectors. Despite the progress in deep architectures for point clouds, previous methods~\cite{pointnet,continuous_conv,point_conv,dynamic_graph_conv,interp_conv} focus purely on utilizing raw point clouds~(for classification~\cite{dataset-pointclouds-shape-classification} or segmentation tasks~\cite{dataset-pointclouds-indoor-scene-segmentation,dataset-pointclouds-shape-segmentation}) instead of fusing them with different sets of sensor information. 

In this section, we introduce an image-guided feature extraction layer for point clouds, called \emph{FusionConv}, which is specialized in sensor fusion-based depth estimation task. Under the known geometric mapping relation between image and point clouds, we design a FusionConv layer that exploits image guidance in several ways. (1)~We adaptively cluster neighbors of each point while considering geometric relation~(3D metric domain) as well as the semantic relation~(2D image domain), which enables us to determine relevant neighbors. (2)~We directly fuse the input point feature with the corresponding image feature via interpolation, which implicitly helps to extract distinctive point features.

\begin{figure}[!t]
    \vspace{+3mm}
    \centering
    \includegraphics[width=1.0\linewidth]{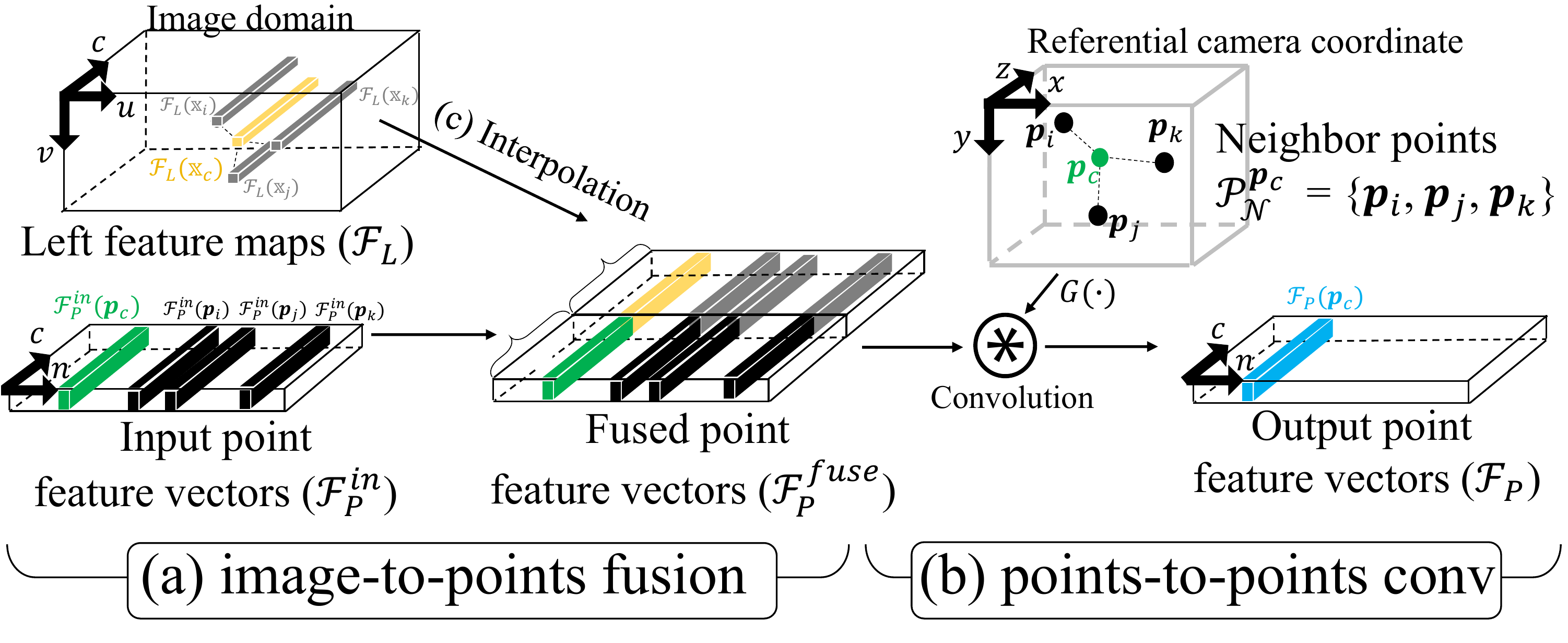}
    \caption{\textbf{FusionConv.} 
    We illustrate the process of FusionConv layer that extracts output point feature vector $\mathcal{F}_{\mathcal{P}}(\textbf{p}_{c})$ from neighboring point clouds $\mathcal{P}^{\textbf{p}_{c}}_{\mathcal{N}}$ of a point $\textbf{p}_{c}$. 
    In (a) image-to-point fusion, interpolation is used to fuse $\mathcal{F}_{L}$ and $\mathcal{F}_{\mathcal{P}}^{in}$ and generate the fused point feature vector $\mathcal{F}_{\mathcal{P}}^{fuse}$.
    Then, we operate (b) points-to-points convolution of fused features $\mathcal{F}_{\mathcal{P}}^{fuse}$ and geometric distance $G(\cdot)$ between adjacent points $\mathcal{P}^{\textbf{p}_{c}}_{\mathcal{N}}$ to infer output point feature vector at the point $\textbf{p}_{c}$, denoted as $\mathcal{F}_{\mathcal{P}}(\textbf{p}_{c})$.
    Note that the overall flow is processed after the clustering stage {($\mathcal{P}^{\textbf{p}_{c}}_{\mathcal{N}} = \{ \textbf{p}_{i}$, $\textbf{p}_{j}$, $\textbf{p}_{k} \} $).}
    }
    \label{fig:fig_point}
\end{figure}

The proposed FusionConv takes the input feature map of the left image~$\mathcal{F}_{\mathcal{L}}$ and the input point feature vectors~$\mathcal{F}_{\mathcal{P}}^{in} \in \mathbb{R}^{C \times N}$ (extracted from raw point clouds~$\mathcal{P}\in \mathbb{R}^{3 \times N}$) and estimates the output point feature vectors~${\mathcal{F}}_{\mathcal{P}} \in \mathbb{R}^{C \times N}$ by fusing the two different features while taking account of relevant neighbors (see \Fref{fig:fig_point}). For simplicity, let $\textbf{p}\in\mathcal{P}$ and $\textbf{x}$ be a 3D point and its projected image point by $\mathbf{K}$, respectively. We then denote the corresponding 3D feature vector and 2D image feature vector as $\mathcal{F}_{\mathcal{P}}(\textbf{p}) \in \mathbb{R}^{C \times 1}$ and $\mathcal{F}_{\mathcal{L}}(\textbf{x}) \in \mathbb{R}^{C \times 1}$, respectively.

Through the mapping relation, the FusionConv layer clusters the neighboring point feature vectors to aggregate the local response in point feature vectors. While many approaches~\cite{continuous_conv,point_conv,dynamic_graph_conv,interp_conv} cluster point clouds $\mathcal{P}$ only in a metric space, we need to cluster the neighboring point clouds following the voxel alignment in the volume $\mathcal{V}$, which is also aligned in the referential camera view. Specifically, for each point $\textbf{p}$ in $\mathcal{P}$, we dynamically cluster neighboring point clouds $\mathcal{P}_{\mathcal{N}}$ 
that reside within the pre-defined size of the voxel window. For instance in~\Fref{fig:fig_point}, there are three neighboring points $\textbf{p}_{i}$, $\textbf{p}_{j}$ and $\textbf{p}_{k}$ within the voxel window whose center indicates the point $\textbf{p}_{c}$. From these three neighbor points and the center point $\textbf{p}_{c}$ itself, FusionConv layer is able to calculate the local response of the point feature vectors at the point $\textbf{p}_{c}$ in alignment with the fusion volume $\mathcal{V}$.

The following process of the FusionConv layer consists of image-to-points fusion and points-to-points aggregation. For the fusion process, we first interpolate $\mathcal{F}_{L}$ into $\mathcal{F}_{\mathcal{P}}$ following the projection mapping relation to obtain the fused point feature vectors $\mathcal{F}_{\mathcal{P}}^{fuse}$. These fused features $\mathcal{F}_{\mathcal{P}}^{fuse}$ have the same number of points as $N$ but they have an extended length of channels containing both $\mathcal{F}_{L}$ and $\mathcal{F}_{\mathcal{P}}^{in}$. After the image-to-points fusion, we convolve the fused point feature vectors $\mathcal{F}_{\mathcal{P}}^{fuse}$ and the geometric distance between the neighboring points $\mathcal{P}_{\mathcal{N}}$ to aggregate the local response, called points~($\mathcal{P}^{\mathcal{N}}$)-to-points~({$\mathcal{F}_{\mathcal{P}}^{fuse}$}) convolution. For instance in~\Fref{fig:fig_point}, the weighted geometric distance $G(\cdot)$ from the center point $\textbf{p}_{c} {=} [x_{c}, y_{c}, z_{c}]^\top$ to its neighboring point $\textbf{p}_{i} {=} [x_i, y_i, z_i]^\top$ is calculated as:
\begin{equation}
    G(\textbf{p}_{c} {-} \textbf{p}_{i}) {=} G(\Delta \textbf{p}) {=} { A_{0} + A_{1}{\cdot}\Delta x + A_{2}{\cdot}\Delta y + A_{3}{\cdot}\Delta z,}
    \label{eq:geometric distance}
\end{equation}
where $A_{0}$, $A_{1}$, $A_{2}$ and $A_{3}$ are learnable weights and we define $\Delta \textbf{p} {=}$ $[\Delta x, \Delta y, \Delta z]^\top$ as $\Delta x {=} x_c {-} x_i$, $\Delta y {=} y_c {-} y_i$, and $\Delta z {=} z_c {-} z_i$.
This weighted geometric distance $G(\cdot)$ becomes the weight of the convolution with the fused point feature vector $\mathcal{F}_{\mathcal{P}}^{fuse}$ to compute the local response at point $\textbf{p}_{c}$ as: 
\begin{equation}
    \mathcal{F}_{\mathcal{P}}(\textbf{p}_{c}) =
    \frac{1}{| \mathcal{P}_{\mathcal{N}}^{{\mathbf{p}_c}} | }
    \sum_{\textbf{p}_{i}\in \mathcal{P}_{\mathcal{N}}^{{\mathbf{p}_c}}} 
    \mathcal{F}^{fuse}_{\mathcal{P}}(\textbf{p}_{i}) \cdot G(\textbf{p}_{c} - \textbf{p}_{i}),
\end{equation}
where $| \mathcal{P}_{\mathcal{N}}^{\mathbf{p}_c} |$ is the number of neighboring points near the point $\textbf{p}_{c}$. Thus, this is the way of imposing different weights $G(\cdot)$ on the neighboring point features $\mathcal{F}_{\mathcal{P}}^{in}$ and left feature maps $\mathcal{F}_{L}$ when extracting the center point feature vectors $\mathcal{F}_{\mathcal{P}}(\textbf{p}_{c})$. To fully compute the output feature vectors from all point clouds, the FusionConv layer iteratively calculates the output point feature vectors $\mathcal{F}_{\mathcal{P}}$ as:
\begin{equation}
    \mathcal{F}_{\mathcal{P}} =
    \begin{bmatrix}
        \mathcal{F}_{\mathcal{P}}(\textbf{p}_{1}), \ &
        \mathcal{F}_{\mathcal{P}}(\textbf{p}_{2}), \ & 
        \cdots{,} \ &
        \mathcal{F}_{\mathcal{P}}(\textbf{p}_{N})
    \end{bmatrix}.
\end{equation}

Finally, we extract the output point feature vector $\mathcal{F}_{\mathcal{P}}~\in~\mathbb{R}^{C \times N}$ from raw point clouds $\mathcal{P}$ and left feature maps $\mathcal{F}_{L}$. This point feature vector becomes embedded into the modified fusion volume $\mathcal{V}_{\mathcal{F}_{\mathcal{P}}} \in \mathbb{R}^{3C \times D \times H \times W}$ as in~\Fref{fig:fig_arch}. The embedded location of $\mathcal{F}_{\mathcal{P}}$ is identical to the corresponding spatial locations of raw point clouds $\mathcal{P}$ within the extended channel-wise voxels $2C+1 \ (\mathcal{V}) \rightarrow 3C \ (\mathcal{V}_{\mathcal{F}_{\mathcal{P}}})$ to maintain the metric accuracy from raw point clouds $\mathcal{P}$. With this fusion volume $\mathcal{V}_{\mathcal{F}_{\mathcal{P}}}$, we can fuse the two different modalities to compute the subpixel matching cost by the following 3D convolution layers as in~\Fref{fig:fig_arch}.


\begin{table*}[t]
    \vspace{+3mm}
    \centering
    \caption{\textbf{Quantitative results of depth estimation networks in KITTI Completion validation benchmark.} \hspace{\textwidth} * represents the reproduced results.
    }
    \vspace{-2mm}
    
    \resizebox{0.7\linewidth}{!}{
        \begin{tabular}{|c|c|c|c|c|c|}
        \hline
        \multirow{3}{*}{Method} & \multicolumn{1}{c|}{\multirow{3}{*}{Modality}} & \multicolumn{4}{c|}{Depth Evaluation} \\ 
        
         & \multicolumn{1}{c|}{} & \multicolumn{4}{c|}{(Lower the better)} \\ \cline{3-6} 
         & \multicolumn{1}{c|}{} & RMSE (mm) & MAE (mm)& iRMSE (1/km) & iMAE (1/km) \\ \hline
        
        
        PSMnet*~\cite{psmnet}
        & Stereo         & 884 & 332 & 1.649 & 0.999  \\ \hline\hline
        
        Sparse2Dense*~\cite{sparse-to-dense}
        & Mono + LiDAR   & 840.0 & - & - & - \\ \hline
        
        Guidenet~\cite{guidenet}
        & Mono + LiDAR   & 777.78 & 221.59 & 2.39 & 1.00 \\ \hline
        
        NLSPN~\cite{jinsunpark}
        & Mono + LiDAR   & 771.8 & 197.3 & 2.0 & 0.8 \\ \hline
        
        CSPN++~\cite{cspn++}
        & Mono + LiDAR   & 725.43 & 207.88 & - & - \\ \hline\hline
        
        Park~\etal~\cite{stereolidar_00} 
        & Stereo + LiDAR & 2021.2 & 500.5 & 3.39 & 1.38 \\ \hline
        
        LiStereo~\cite{stereolidar_01} 
        & Stereo + LiDAR & 832.16 & 283.91 & 2.19 & 1.10 \\ \hline
        
        CCVN~\cite{stereolidar_norm_costV_ccvn} 
        & Stereo + LiDAR & 749.3 & 252.5 & \textbf{1.3968} & \textbf{0.8069} \\ \hline\hline
        
        Ours 
        & Stereo + LiDAR & \textbf{636.2} & \textbf{205.1} & 1.8721 & 0.9870 \\ \hline
        
        \end{tabular}
    }
    \vspace{-1mm}
    \label{table:metric-kitti}
\end{table*}
\begin{table}[!t]
\centering
\caption{\textbf{Quantitative results of depth estimation networks in the Virtual-KITTI 2.0 dataset.} \hspace{\textwidth}
* represents the reproduced results. S, M+L and S+L represent stereo cameras, monocular camera with a LiDAR and stereo cameras with a LiDAR respectively.
}
\vspace{-2mm}
\resizebox{1.0\linewidth}{!}{
\begin{tabular}{|c|c|c|c|c|c|}
\hline
\multirow{4}{*}{Method} & \multicolumn{1}{c|}{\multirow{4}{*}{Modality}} & \multicolumn{4}{c|}{Depth Evaluation} \\

& \multicolumn{1}{c|}{} & \multicolumn{4}{c|}{(Lower the better)} \\ \cline{3-6} 
    
& \multicolumn{1}{c|}{} & RMSE & MAE & iRMSE & iMAE \\ 
& \multicolumn{1}{c|}{} & (mm) & (mm)& (1/km) & (1/km) \\ \hline


\text{ } PSMnet~\cite{psmnet}* & S & 5728 & 2235  & 9.805 & 4.380 \\ \hline

\text{ } Sparse2Dense~\cite{sparse-to-dense}* & M+L &{3357}$\pm$8 & 1336$\pm$2.0 & 12.136$\pm$0.045 & 6.243$\pm$0.013
\\ \hline

\text{ } CCVN~\cite{stereolidar_norm_costV_ccvn}* & S+L & 3726.83$\pm$10.83 & 915.6$\pm$0.4 & 8.814$\pm$0.019 & \textbf{2.456}$\pm$\textbf{0.004} \\ \hline\hline

Ours & S+L & \textbf{3217.16}$\pm$\textbf{1.84} & \textbf{712}$\pm$\textbf{2.0} & \textbf{7.168}$\pm$\textbf{0.048} & 2.694$\pm$0.011 \\ \hline

\end{tabular}
}
\label{table:metric-virtual-kitti}
\end{table}

\section{Depth map regression}
\label{sec:Depth regression}

After we fuse features from the two modalities at the fusion volume $\mathcal{V}_{\mathcal{F}_{\mathcal{P}}}$, we propagate the point features to the overall volume and compute the matching cost through the stacked hourglass networks~\cite{hourglass,psmnet} to regress the depth map. Following~\cite{psmnet}, we perform a cost aggregation process by aggregating the fusion volume along the depth dimension as well as the spatial dimension. In our stacked hourglass networks, the networks consist of three encoder-decoder networks that sequentially refine the cost aggregation via intermediate loss~\cite{psmnet,hourglass,ganet}. After aggregation, the cost aggregation reduces the channels of $\mathcal{V}_{\mathcal{F}_{\mathcal{P}}}$ into a 3D structure $\mathcal{A} \in \mathbb{R}^{D \times H \times W}$. From $\mathcal{A}$, we can estimate the depth value $\Tilde{z}_{u, v}$ at pixel $(u, v)$ as: \vspace{-1mm}
\begin{equation}
    \Tilde{z}_{u, v} = \sum_{d=0}^{D-1} \frac{d}{D-1} \cdot z_{max} \cdot \sigma ({\mathbf{a}}_{u,v}^{d}),
\vspace{-1mm}
\end{equation}
where $z_{max}$ is a hyper-parameter defining the maximum range of depth estimation, $\sigma(\cdot)$ represents the softmax operation, and ${\mathbf{a}}_{u,v}^{d}$ is the $d$-th value of the cost aggregation vector ${\mathbf{a}}_{u,v} \in \mathbb{R}^{D\times 1} \text{ at } (u, v)$. For our experiments, we set the hyper-parameters as $D=48$ and $z_{max}=100$. Specifically, we compute the depth loss~$\mathcal{L}_{depth}$ from the estimated depth map~$\tilde{\mathcal{Z}}$ and the true depth map~$\mathcal{Z}$ as follows: \vspace{-1mm}
\begin{equation}
    \mathcal{L}_{depth} = \frac{1}{M}\sum_{u}\sum_{v} smooth_{L_{1}}({z}_{u,v} - \tilde{z}_{u,v}),
    \vspace{-1mm}
    \label{equation:disp-loss}
\end{equation}
where $M$ is the number of valid pixels in $\mathcal{Z}$ for the normalizing factor, $\tilde{z}_{u,v}$ is the value of the predicted depth map $\widetilde{\mathcal{Z}}$ at pixel location $(u, v)$, and $smooth_{L_{1}}(\cdot)$ is the smooth L1 loss function used to compute the loss~\cite{gcnet,psmnet}. Finally, the total loss $\mathcal{L}_{total}$ of our network is computed from the three different intermediate results of depth maps from the three stacked hourglass networks~\cite{psmnet,hourglass} as below:
\begin{equation}
    \mathcal{L}_{total} = \sum_{i=1}^{3} w_{i} \cdot \mathcal{L}_{depth}^{i},
\end{equation}
where $w_i$ is the weight of the $i$-th depth loss $\mathcal{L}_{depth}^{i}$ (we set the weight to $w_{1}{=}0.5$, $w_{2}{=}0.7$ and $w_{3}{=}1.0$). During the evaluation, we only consider the predicted depth map at the last network, as illustrated in~\Fref{fig:fig_arch}. 


\section{Experimental Evaluation}
\label{sec:Experimental Evaluation}
In this section, we describe the implementation details of our network. Our network is trained on two independent datasets, the KITTI dataset~\cite{kitti-completion} and the Virtual-KITTI dataset~\cite{virtual-kitti}. We separately evaluate the accuracy of the depth for each dataset against existing techniques. Additionally, we conduct an ablation study to validate each dominant component of our method and to contrast the early fusion~\cite{stereolidar_norm_costV_ccvn} with our intermediate fusion at the fusion volume $\mathcal{V}_{\mathcal{F}_{\mathcal{P}}}$.

\vspace{-1mm}
\subsection{Architecture}
\label{subsec:Architecture}
Our network consists of three stages: the feature extraction stage, volume aggregation stage, and depth regression stage, as depicted in~\Fref{fig:fig_arch}. In the feature extraction stage, we follow the architecture of image feature extraction layers as in Chang and Chen~\cite{psmnet}, but extract the intermediate left feature map to operate image-to-points fusion in FusionConv layers. We use three FusionConv layers to infer the point feature vectors. Then, the extracted features from input modalities are embedded in the fusion volume as explained in~Secs.~\textcolor{blue}{\RNum{4}} and \textcolor{blue}{\RNum{5}}. We aggregate the volume through the three stacked hourglass networks and finally regress the depth map as described in~\Sref{sec:Depth regression}. 

\begin{figure*}[!t]
\vspace{+3mm}
\centering
\includegraphics[width=0.90\linewidth]{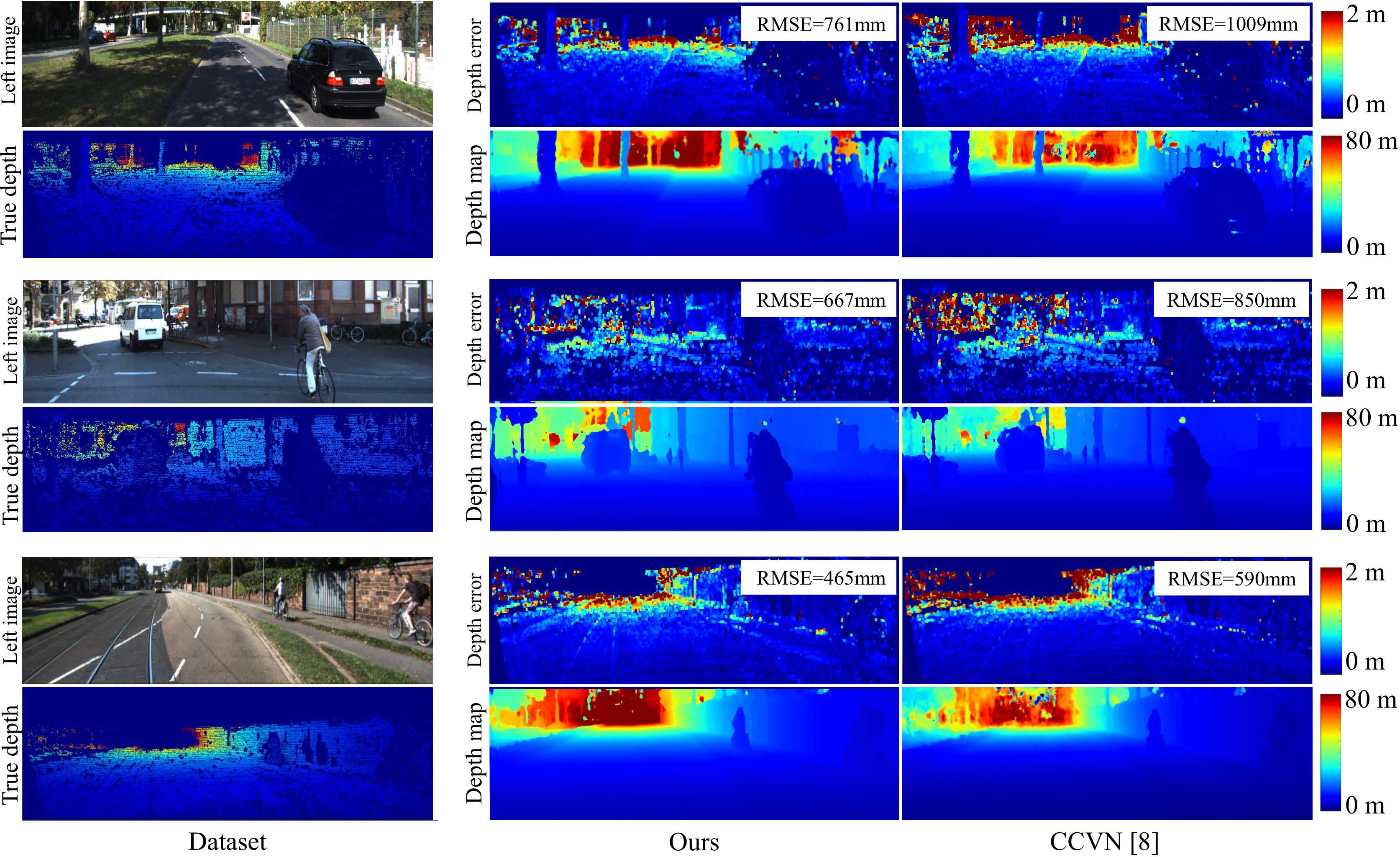}
\vspace{-3mm}
\caption{\textbf{Qualitative results on the KITTI dataset.} 
We visualize depth maps and depth errors from ours and the recent stereo-Lidar method (CCVN~\cite{stereolidar_norm_costV_ccvn}) in three different cases. 
We also include the depth metric RMSE (lower the better).
}
\label{fig:fig_result}
\vspace{-2mm}
\end{figure*}

\vspace{-1mm}
\subsection{Datasets and training schemes.}
\label{subsec:Datasets and training schemes.}

\noindent \textbf{KITTI dataset.} \
The KITTI Raw benchmark~\cite{kitti} provides sequential stereo images and LiDAR point clouds under different road environments. Within the benchmark, KITTI completion benchmark~\cite{kitti-completion} provides the true depth maps and the corresponding sensor data, consisting of 42,949 training samples and 1,000 validation samples. Given the input sensory data, we train our network by setting the learning rate as 0.001 for 5 epochs and as 0.0001 for 30 epochs. We use three NVIDIA 1080-Ti GPUs for training, and the batch size is 9. The entire training scheme takes three days and the inference speed of our network during the test is 0.71 FPS (1.40 sec per frame). We use random-crop augmentation during the training phase. For training, the size of the cropped images is $256 \times 512$.  We also crop point clouds $\mathcal{P}$ that reside within the cropped left images. Usually, there are $\sim$5K point clouds in the training phase, but it can vary depending on the location of the cropped area within images. For testing, we fully utilize the original shape of images and raw point clouds ($\sim$25K points per image) without any augmentation or filtering.

\noindent \textbf{Virtual KITTI 2.0 dataset.} \
Virtual KITTI 2.0~\cite{virtual-kitti} is a recently published dataset that provides much more realistic images than the previous version of the Virtual-KITTI 1.3.1~\cite{virtual-kitti-cvpr2016}. The merits of the synthetic environment include access to dense ground truth at the farther areas, while the KITTI completion benchmark~\cite{kitti-completion} provides relatively sparse ground truth. There are five scenes in this dataset and each scene contains ten different scenarios, such as rain, sunset, \etc. We set the two scenes (Scene01, Scene02) for training the network and the other scenes are exploited for evaluating the accuracy of depth maps. For each scene, we take the only scenario (15-deg-left) for training and evaluation. In total, there are 680 images in the training set and 1,446 images in the test set. Though the raw point cloud data is not provided in this dataset, we randomly sample the ground truth depth pixels and regard the pixels as the point clouds. We select the same number of selected point clouds as for the KITTI dataset (\ie,~5K points for training and 25K points for testing). 
%
%
Given pre-trained weights from the KITTI dataset, 
we fine-tune the network for 5K iterations under the identical augmentation methods as we adopt for the KITTI dataset.

\noindent \textbf{Metrics.} \
We evaluate the quality of the estimated depth by our network. We follow the metric scheme proposed by Eigen~\etal~\cite{eigen_mono_depth} which is identical to the official KITTI depth completion benchmark~\cite{kitti-completion}, \ie, RMSE, MAE, iRMSE, iMAE, where RMSE and MAE are the target metrics or evaluating the metric distance. These metric formulations are equivalently applied to measure the performance on the Virtual-KITTI dataset~\cite{virtual-kitti}. Since the Virtual-KITTI dataset does not provide raw point clouds data, we evaluate the depth metric by 5 times of repetitive sampling of the ground truth depth pixels. The resulting metric in~\Tref{table:metric-virtual-kitti} is the average over the samples. 
Note that we re-train the network by accessing the open-source code of the target method and denote the re-evaluated results by a `*' (\eg, PSMnet*) as in~Tables~\textcolor{blue}{\RNum{1}} and~\textcolor{blue}{\RNum{2}}.

\begin{figure*}[!t]
\vspace{+3mm}
\centering
\includegraphics[width=0.90\linewidth]{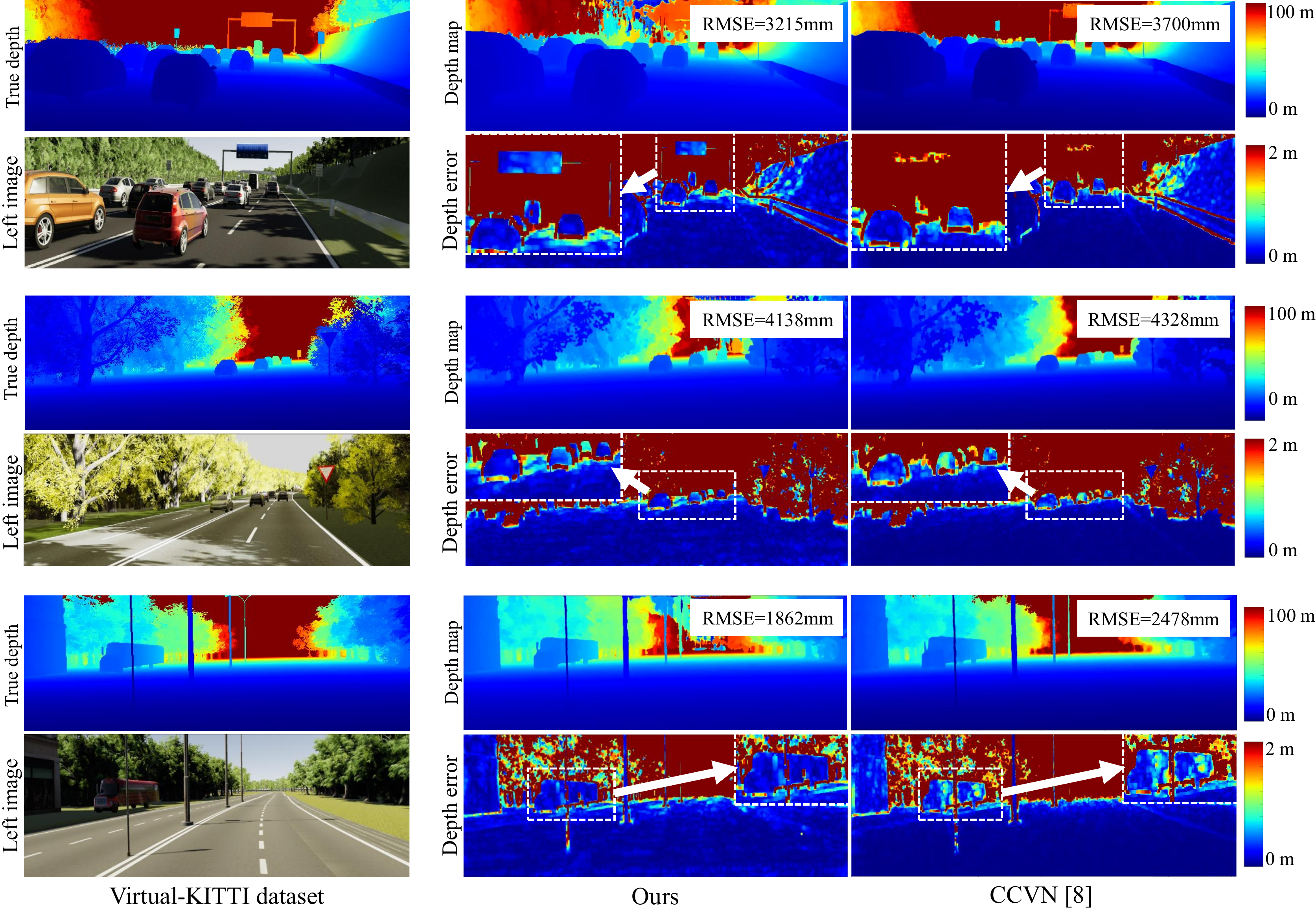}
\vspace{-3mm}
\caption{\textbf{Qualitative results on the Virtual-KITTI 2.0 dataset.} 
We evaluate the estimated depth from both our network and the recent Stereo-LiDAR fusion network by Wang~\etal~\cite{stereolidar_norm_costV_ccvn}.
This synthetic data covers the wide range of depth upto $655m$,
but we clamp true depth maps and estimated depth maps upto $100m$ during the evaluation, as in Table~\textcolor{blue}{\RNum{2}}. 
For the detailed visualization, we crop and enlarge the part of depth error maps in each frame.
Mainly, the cropped images correspond to the farther area to validate our long-range depth estimation.
}
\label{fig:fig_result_virtual_kitti}
\vspace{-2mm}
\end{figure*}

\noindent \textbf{Comparison.} \
We evaluate our network against existing methods~\cite{stereolidar_01,stereolidar_00,stereolidar_norm_costV_ccvn} for the KITTI completion validation benchmark (\Tref{table:metric-kitti}) and the Virtual-KITTI 2.0 (\Tref{table:metric-virtual-kitti}). We also include the performance of other depth estimation networks, such as stereo matching methods~\cite{psmnet,gcnet} and depth completion studies~\cite{sparse-to-dense,guidenet,jinsunpark,cspn++}. Among the existing methods, our method achieves state-of-the-art depth performance in RMSE and MAE as in Tables~\textcolor{blue}{\RNum{1}}~and~\textcolor{blue}{\RNum{2}}. These results suggest that our method shows higher accuracy in the estimation of the depth at a distant area. This is consistent with our intention to create a metric-linear volumetric design. To further verify the strength of our method, we provide the qualitative results for the KITTI dataset and the Virtual-KITTI dataset in Figs.~\textcolor{blue}{3}~and~\textcolor{blue}{4}. We attribute this improvement to the fusion at the fusion volume that simultaneously computes matching cost between features of the two input modalities. More quantitative results and information about the inference time are available in the supplementary material.

\vspace{-1mm}
\section{Ablation study}
\label{sec:Ablation study}
In this section, we extensively investigate our neural modules: the fusion volume and the FusionConv layer. 
The following experiments are conducted on the KITTI dataset~\cite{kitti-completion}.


\begin{table}[!t]
    \centering
    \caption{\textbf{Ablation study of the type of volume for depth estimation.} We differentiate the type of volume as cost volume~\cite{gcnet} (\ie,~CostV), depth volume (\ie,~DepthV) by You~\etal~\cite{pseudo-lidar++}, and our fusion volume $\mathcal{V}$ (\ie,~ FusionV). 
    Note that depth volume (\ie,~DepthV) by You~\etal~\cite{pseudo-lidar++} is a transformed volume from the cost volume, but our method directly interpolates sensor data into our volume $\mathcal{V}$.
    We evaluate each method on the KITTI completion validation benchmark~\cite{kitti-completion}. 
    }
    \vspace{-2mm}
    \resizebox{1.0\linewidth}{!}{
    \begin{tabular}{|c|c|c|c|c|c|c|c|}
    \hline
    
    \multirow{3}{*}{} & \multicolumn{3}{c|}{Preserve (\checkmark)} & \multicolumn{4}{c|}{Depth Evaluation} \\ \cline{2-4} 
    
    & \multicolumn{3}{c|}{Type of Volume} & \multicolumn{4}{c|}{(Lower the better)} \\ \cline{2-8} 
    
    & CostV        & DepthV                & FusionV & RMSE & MAE  & iRMSE  & iMAE \\ 
    & \cite{gcnet} & \cite{pseudo-lidar++} & (ours)  & (mm) & (mm) & (1/km) & (1/km) \\ \hline
    
    1 & $\checkmark$ &              &              & 884            & 332            & \textbf{1.649} & 0.999 \\ \hline
    
    2 &              & $\checkmark$ &              & 797            & 286            & 2.046  & 1.070  \\ \hline
    
    3 &              &              & $\checkmark$ & \textbf{636.2} & \textbf{205.1} & 1.8721 & \textbf{0.9870} \\ \hline
    
    \end{tabular}
    }
    \label{table:ablation-type-of-volume}
    \vspace{-4mm}
\end{table}

\noindent \textbf{Fusion volume.} \
We compare our fusion volume with other volume representations, such as cost volume~\cite{gcnet} and depth volume~\cite{pseudo-lidar++} as shown in~\Tref{table:ablation-type-of-volume}. Note that the depth volume~\cite{pseudo-lidar++} is built via cost volume, but our fusion volume directly encodes point feature vectors $\mathcal{F}_{\mathcal{P}}$ into the metric-aware voxels. For a fair comparison, we evaluate each type of volume under identical conditions, such as deep architectures (\eg,~FusionConv) and input sensory data (\eg,~stereo-lidar fusion). In~\Tref{table:ablation-type-of-volume}, our fusion volume shows the highest accuracy among different voxel representations. We deduce the reason that our volume can directly encode the 3D metric point into the volume, while the other voxel representation~\cite{pseudo-lidar++} is constructed via cost volume, which can lose metric information during the transformation.

\begin{table}[!t]
    \centering
    \caption{\textbf{Ablation study of type of fusion for depth estimation.}
    Each method embeds different information into the fusion volume, such as raw point cloud $\mathcal{P}$, point feature vectors from multilayer perceptron (MLP) by Qi~\etal~\cite{pointnet} and point feature vectors from our FusionConv layer.
    }
    \vspace{-2mm}
    \resizebox{1.0\linewidth}{!}{
    \begin{tabular}{|c|c|c|c|c|c|c|c|}
    \hline
    
    \multirow{4}{*}{} & \multicolumn{3}{c|}{Preserve (\checkmark)} & \multicolumn{4}{c|}{Depth Evaluation} \\ \cline{2-4}
    
    & \multicolumn{3}{c|}{Type of point network} & \multicolumn{4}{c|}{(Lower the better)} \\ \cline{2-8}
    
    & Raw    & MLP & FusionConv & RMSE & MAE & iRMSE & iMAE \\ 
    
    & \ points \ & \cite{pointnet} & (ours) & (mm) & (mm) & (1/km) & (1/km) \\ \hline
    
    4 & $\checkmark$ &              &              & 669 & 226 & 2.169 & 1.120 \\ \hline
    
    5 &             & $\checkmark$ &              & 652 & 212 & 2.099 & 1.066 \\ \hline
    
    
    6 &              &              & $\checkmark$ & \textbf{636.2} & \textbf{205.1} & \textbf{1.8721} & \textbf{0.9870} \\ \hline
    
    \end{tabular}
    }
    \label{table:ablation-fusion-conv}
    \vspace{-4mm}
\end{table}
\begin{table}[!t]
    \centering
    \caption{\textbf{Ablation study of different levels of fusion}. Early fusion takes pre-processing steps to project point clouds into image domain for fusion, while intermediate fusion proposes fusion at feature space, \eg, fusion volume.
    }
    \vspace{-2mm}
    \resizebox{1.0\linewidth}{!}{
    \begin{tabular}{|c|c|c|c|c|c|c|c|c|}
    \hline
    
    \multirow{4}{*}{} & \multicolumn{3}{c|}{Preserve (\checkmark)} & \multicolumn{4}{c|}{Depth Evaluation} \\  \cline{2-4} 
    
    & \multicolumn{2}{c|}{Level of fusion} & \multirow{3}{*}{Volume} & \multicolumn{4}{c|}{(Lower the better)} \\ \cline{2-3} \cline{5-8} 
    
    & Early & Intermediate & & RMSE & MAE & iRMSE & iMAE \\ 
    
    & fusion & fusion \small{(ours)} & & (mm) & (mm) & (1/km) & (1/km) \\ \hline
    
    7 & $\checkmark$ &              & CostV & 744            & 249            & 2.026  & 1.022 \\ \hline
    
    8 & $\checkmark$ &              & FusionV & 650            & 215            & 1.912  & \textbf{0.964} \\ \hline
    
    9 &              & $\checkmark$ & FusionV & \textbf{636.2} & \textbf{205.1} & \textbf{1.8721} & 0.9870 \\ \hline
    
    \end{tabular}
    }
    \label{table:ablation-early-fusion}
    \vspace{-6mm}
\end{table}

\noindent \textbf{FusionConv.} \
In~\Tref{table:ablation-fusion-conv}, we decompose FusionConv into three factors: convolution, cluster, and fusion, to analyze each component in the proposed layers. As a baseline (Method~\textcolor{blue}{4} in~\Tref{table:ablation-fusion-conv}), we embed the raw point clouds $\mathcal{P}$ into the volume $\mathcal{V}_{\mathcal{P}}$ as described in~\Sref{sec:Volumetric Propagation}. Moreover, we set another baseline network as a multilayer perceptron layer (\ie,~fully-connected layer) for point feature extraction~\cite{pointnet} (Method~\textcolor{blue}{5}), which does not infer point feature vectors $\mathcal{F}_{\mathcal{P}}$ via clustering. Though the quality of depth from the two baseline methods outperforms the previous method~\cite{stereolidar_norm_costV_ccvn}, the convolution operation among the neighboring points~$\mathcal{P}_{\mathcal{N}}$, as in FusionConv, further increases the accuracy of the depth estimation (Method~\textcolor{blue}{6}). This confirms that our clustering strategy and image-to-point fusion are effective for the sensor fusion-based depth estimation task. Finally, our FusionConv layer (Method~\textcolor{blue}{6}) shows the highest metric performance.

\noindent \textbf{Early fusion.} \
Our method proposes fusion in the feature space, called intermediate fusion, through the fusion volume. However, previous methods~\cite{stereolidar_01,stereolidar_00,stereolidar_norm_costV_ccvn} internally use the pre-processing stage to project raw point clouds $\mathcal{P}$ into the image domain to concatenate them with RGB images, called early fusion. With this ablation study, we validate the performance gap between early fusion and our intermediate fusion to prove the effectiveness of fusion in a 3D metric space. For a fair comparison with different levels of fusion, we use similar architectures with different types of volume as listed in~\Tref{table:ablation-early-fusion}. For early fusion, we do not embed point clouds into the volume but undergo the pre-processing stage as previous works proposed~\cite{stereolidar_01,stereolidar_00,stereolidar_norm_costV_ccvn}. Method~\textcolor{blue}{9} represents our method. In~\Tref{table:ablation-early-fusion}, it reveals that intermediate fusion in a 3D voxel space (Methods~\textcolor{blue}{8},~\textcolor{blue}{9}) shows better results than the early fusion approach (Method~\textcolor{blue}{7}). We deduce that our fusion scheme takes into consideration the spatial connectivity of point clouds and stereo images for seamless fusion and obtains long-range depth maps.

\vspace{-1mm}
\section{Conclusion}
\label{sec:Conclusion}
In this paper, we propose a volumetric propagation network for stereo-LiDAR fusion and perform the long-range depth estimation. To this end, we design two dominant modules, fusion volume and FusionConv, to facilitate the fusion in a unified volumetric space. Within the fusion volume, we formulate the fusion as volumetric propagation by considering the spatial connectivity of sparse point features and densely-ordered stereo images features. Our method demonstrates state-of-the-art performance on the KITTI and the Virtual-KITTI datasets and delivers a message about the geometric-aware stereo-LiDAR fusion.


\bibliographystyle{IEEEtran}
\bibliography{egbib}

\newpage

\renewcommand*{\thesection}{\Alph{section}}

\vspace{2mm}
{\LARGE \ \ \ \ \  \textbf{Supplementary Material}}
\vspace{2mm}

\setcounter{section}{0}
\setcounter{figure}{0}
\setcounter{table}{0}

\section{Overview}
This document provides additional information that we did not fully cover in the main paper, \textbf{\textit{Volumetric Propagation Network: Stereo-LiDAR Fusion for Long-Range Depth Estimation}}. All references are consistent with the manuscript. 

\section{Volumetric propagation}
\label{supp-sec:Propagation and triangulation}
For the fusion of different sets of sensory information, the recent stereo-LiDAR fusion method (CCVN~\cite{stereolidar_norm_costV_ccvn}) introduces input fusion (\ie, stereo RGB-D input) as well as cost volume normalization conditioned by sparse disparity maps. Typically, this conditional cost volume normalization mainly affects the 3D volumetric aggregation and looks closer to the idea of our volumetric propagation.

\begin{figure}[!t]
\centering
\includegraphics[width=0.95\linewidth]{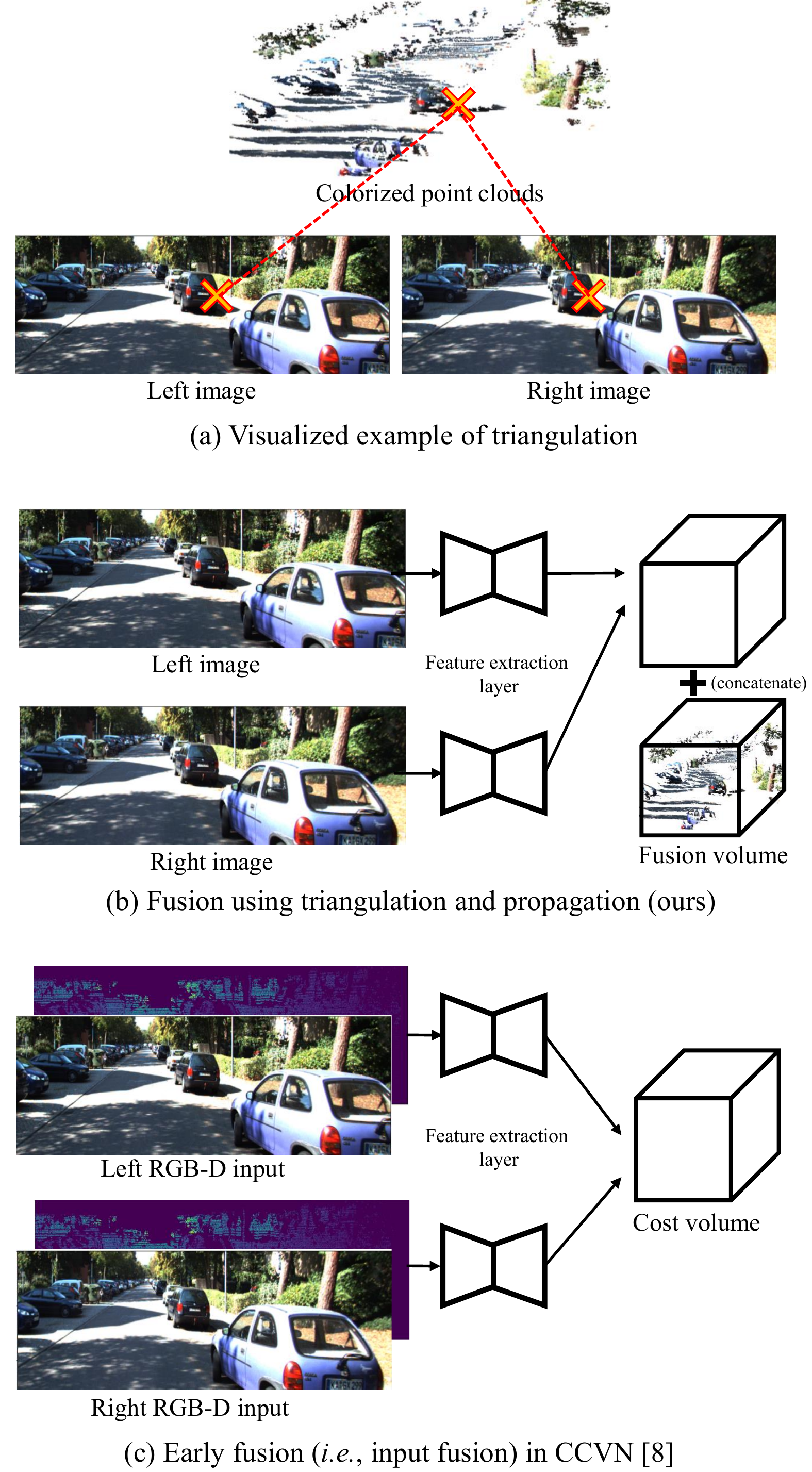}
\caption{\textbf{Examples of fusion mechanism.}
(\textbf{a}) We provide an example of triangulation in stereo images and colorized point clouds. Note that the orange marks~($\boldsymbol{\times}$) represent the 2D-3D correspondence in stereo images and point clouds. 
(\textbf{b}) By embedding point clouds into the fusion volume, we apply the idea of propagation into the fusion of stereo images and point clouds. 
Additionally, (\textbf{c}) we provide conceptual flow of the early fusion (\ie input fusion) proposed by the recent stereo-LiDAR fusion method (CCVN~\cite{stereolidar_norm_costV_ccvn}). 
}
\vspace{-2mm}
\label{fig:triangulation}
\end{figure}

On the other hand, the proposed concept of volumetric propagation has a novel and unique geometric insight. The fundamental reasoning of our method is related to \textbf{triangulation in the 3D space (\ie, our fusion volume)}. By embedding point clouds directly into the 3D volumetric space, we can provide accurate 3D seeds of pixel correspondences between stereo images. 
To further elucidate our point, we provide example visualization in~\Fref{fig:triangulation}.

\section{Quantization error}
\label{supp-sec:Quantization error}
One of the major differences between cost volume~\cite{psmnet,gcnet,ganet} and our fusion volume is the design choice of depth range, which is a matter of different spacing choices along the referential camera ray.

Basically, our intention of using the fusion volume is to seamlessly fuse two different sets of  3D information (one from stereo images and the other one from LiDAR) in a unified 3D space for high-quality and long-range depth estimation. 
In particular, we are interested in improving the depth quality in distant areas, where a pair of stereo images has difficulty in finding the correspondence. To this end, we adopt the metric-scale fusion volume that equally spaces the depth range and has less quantization loss even in the far-away regions.

\begin{figure*}[!t]
\centering
\includegraphics[width=0.86\linewidth]{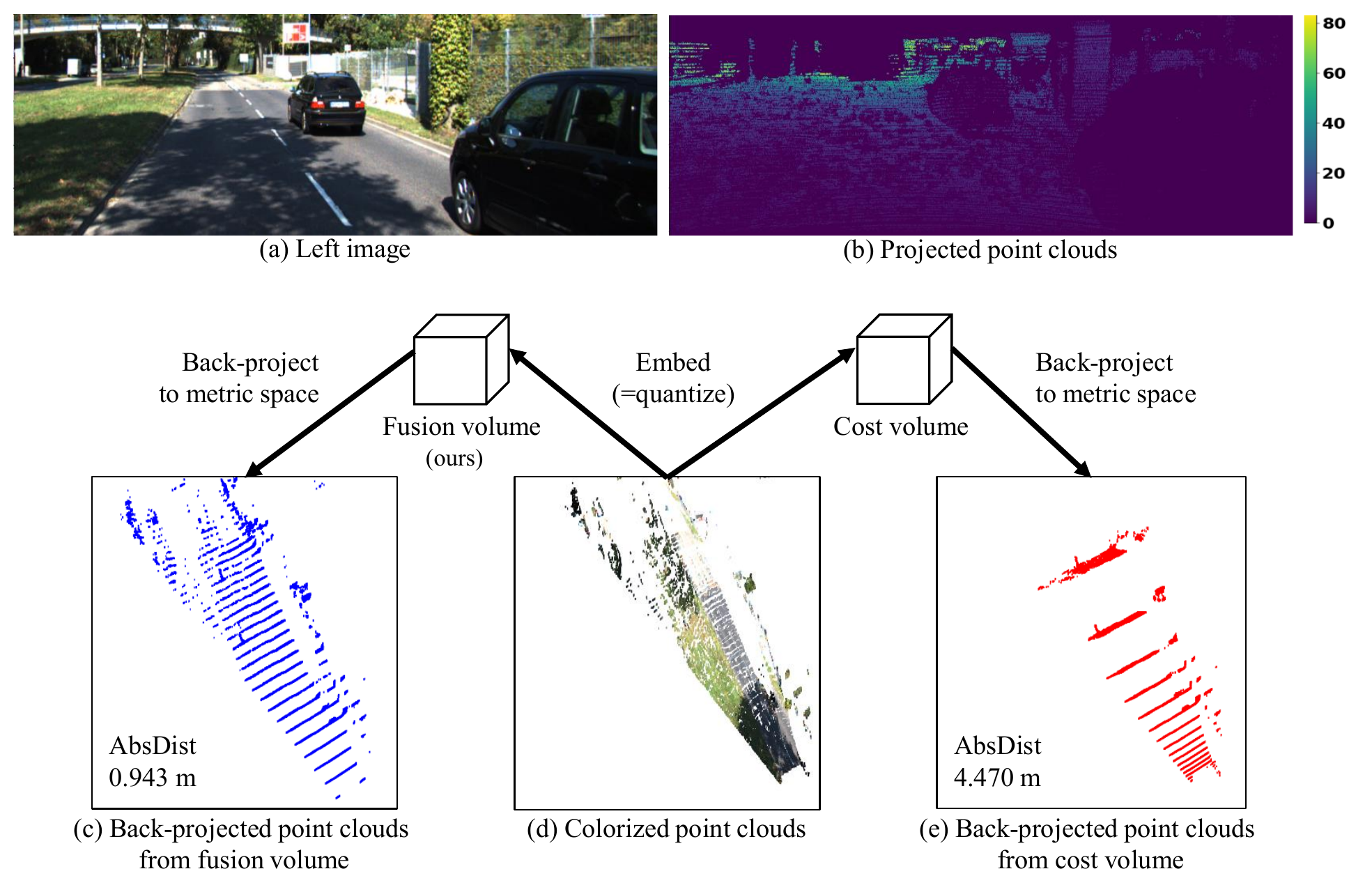}
\vspace{-4mm}
\caption{\textbf{Visualization of quantization loss for embedded point clouds from cost volume and fusion volume.} Given (a) a left image and (b) projected point clouds, we back-project (d) colorized point clouds into the 3D metric space. 
To illustrate the quantization loss for embedded point clouds, we first embed the point clouds into each volume and back-project the points into the 3D metric space. 
Visually, the fusion volume can generally describe the target road scene while the cost volume concentrates on the closer area, which is consistent with our intention to maintain the metric accuracy of point clouds.
Also, we calculate the quantization loss by measuring the absolute distance (\ie, AbsDist) between (d) original point clouds and (c, e) embedded point clouds. 
Our intention of the usage of the word. \textit{quantization loss}, is the metric-scale difference.} 
\label{fig:quantization}
\end{figure*}

For better understanding, we validate our proposed solution by extensive qualitative and quantitative evaluation. We first visualize the quantization loss that comes from embedded points as in~\Fref{fig:quantization}. We then calculate the performance of depth maps in three different ranges: close area ($0m-20m$), middle area ($20-40m$), and far area ($40-80m$) as in~\Tref{table:ablation-range}. The results are consistent with our intention of creating a volumetric design since our method shows outperforms the recent works~\cite{stereolidar_00,stereolidar_01,stereolidar_norm_costV_ccvn}, especially in the far area.


\begin{table*}[!t]
\centering
\caption{\textbf{Ablation study for depth evaluation in different range of distance.} We utilize the KITTI depth completion validation benchmark~\cite{kitti-completion} for evaluation. Note that AbsRel and SqRel represents absolute relative difference and squared relative difference, respectively~\cite{eigen_mono_depth}.}
\vspace{-2mm}
\resizebox{0.86\linewidth}{!}{
\begin{tabular}{|c|c|c|c|c|c|c|c|c|}
\hline
\multirow{3}{*}{ \ \shortstack{Evaluation \\ Depth range} \ } & \multirow{3}{*}{Method} &  \multicolumn{6}{c|}{Depth Evaluation in KITTI} \\ 
& & \multicolumn{6}{c|}{(Lower the better)} \\ \cline{3-8} 
& & AbsRel (m) & SqRel (m) & RMSE (mm) & MAE (mm) & iRMSE (1/km) & iMAE (1/km) \\ \hline
%
%
\multirow{2}{*}{0m $-$ 20m} & CCVN~\cite{stereolidar_norm_costV_ccvn} & \textbf{0.0101} & 0.0078 & 280.8 & \textbf{115.3} & \textbf{1.9214} & \textbf{0.9931} \\ \cline{2-8}
& Ours & 0.0106 & \textbf{0.0037} & \textbf{196.8} & 115.7 & 1.9237 & 1.1057 \\ \hline\hline
\multirow{2}{*}{20m $-$ 40m} & CCVN~\cite{stereolidar_norm_costV_ccvn}  & 0.0196 & 0.0387 & 1032.3 & 557.3 & \textbf{1.4523} & \textbf{0.7240} \\ \cline{2-8}
& Ours & \textbf{0.0112} & \textbf{0.0077} & \textbf{366.5} & \textbf{162.2} & 1.8272 & 1.0029 \\ \hline\hline
%
%
%
\multirow{2}{*}{40m $-$ 80m} & CCVN~\cite{stereolidar_norm_costV_ccvn}  & 0.0373 & 0.1810 & 3151.1 & 2051.6 & \textbf{1.4185} & \textbf{0.7279} \\ \cline{2-8}
& Ours & \textbf{0.0116} & \textbf{0.0136} & \textbf{631.2} & \textbf{205.7} & 1.8018 & 0.9699 \\ \hline
%
 %
\end{tabular}}
\label{table:ablation-range}
\end{table*}


\section{Hyper-parameters}
\label{supp-sec:Hyper-parameters}

In this section, we describe the details of the hyper-parameters that we did not fully describe in the manuscript. Since hyper-parameter tuning can affect the depth performance, we try not to change the specific value but use the proposed values from our baseline method (PSMNet~\cite{psmnet}). For example in the dimension of the channels, given input images with dimension $3 \times 4H \times 4W$, the feature extraction layers infer image feature maps $C({=}32) \times H \times W$. Note that the feature extraction layer is identical to our baseline network~\cite{psmnet}, as stated in~\Sref{sec:Experimental Evaluation}-\textcolor{blue}{A} of the manuscript. Also, the point feature vectors $\mathcal{F}_{\mathcal{P}}$ have the same number of channels $(C{=}32)$. For the composition of the fusion volume, we use $D{=}48$, $z_{max}{=}100$. 
Note that the weights are set as $w_{1}{=}0.5, w_{2}{=}0.7, w_{3}{=}1.0$. The overall hyper-parameters are identical/close to the design choice by our baseline method~\cite{psmnet}. This is for a fair comparison.

\section{Inference time}
\label{supp-sec:Inference time}
This section provides the details of the inference time of each specific stage in our volumetric propagation network, as shown in Tables~\textcolor{blue}{\RNum{2}}~and~\textcolor{blue}{\RNum{3}}. As expected, the volume aggregation stage takes the most inference time, while the other stages take less computation. This is because the volumetric data requires large computation powers. 
On the other hand, the FusionConv shows real-time inference speed in our environment. Because the FusionConv is fully implemented in a GPU-friendly environment using CUDA, the inference time of the FusionConv is less than $27 ms$. Note that the speed is measured during the evaluation and the number of point clouds are less than $25K$. Moreover, we compare the inference speed of our method, recent stereo-LiDAR methods~\cite{stereolidar_00,stereolidar_norm_costV_ccvn}, and baseline methods~\cite{gcnet,psmnet} as in~\Tref{table:inference-time-recent-methods}. For fairness, we use a Core i7 processor and an NVIDIA 1080Ti GPU to measure the speed, which are similar to those of the recent method (CCVN~\cite{stereolidar_norm_costV_ccvn}). Though our method requires more computation power than others, this is not from the implementation of FusionConv but from the volumetric aggregation stage in stacked hourglass networks~\cite{hourglass,psmnet}. 

\begin{table*}[!t]
\centering
\caption{\textbf{Inference speed of each stage of our network.} We split the our network into three stages as illustrated in~Fig.~\textcolor{red}{2} of the manuscript and measure the inference speed of each stage during the test.}
\vspace{-2mm}
\resizebox{1.0\linewidth}{!}{
\begin{tabular}{|c|c|c|c|c|}
\hline
& \multicolumn{3}{c|}{Our overall network} & \multirow{2}{*}{\text{  } Total \text{  }} \\ \cline{2-4} 

& \ Feature extraction stage \ & \ Volume aggregation stage \ & \ Depth regression stage \ & \\ \hline

\ Inference time \ & \multirow{2}{*}{0.027179} & \multirow{2}{*}{1.347444} & \multirow{2}{*}{0.027745} & \multirow{2}{*}{1.40} \\ 

\ (sec) \ & & & & \\ \hline

\end{tabular}
}
\label{table:inference-speed-per-stage}
\vspace{3mm}
\end{table*}
\begin{table*}[!t]
\centering
\caption{\textbf{Inference speed of feature extraction stage.} Our point feature extraction layer, FusionConv, shows fast inference speed in spite of dynamical clustering scheme. The test has been conducted under $\sim$25K point clouds~$\mathcal{P}$ and $384(H) \times 1248(W)$ size of stereo images as input.}
\vspace{-2mm}
\resizebox{1.0\linewidth}{!}{
\begin{tabular}{|c|c|c|c|c|c|}
\hline
& \multicolumn{4}{c|}{Feature extraction} & \multirow{2}{*}{\text{   } Total \text{   }}  \\ \cline{2-5} 
& Image feature extraction & \ 1st Fusion Conv \ & \ 2nd Fusion Conv \ & \ 3rd FusionConv \ & \\ \hline
 

\ Inference time \ & \multirow{2}{*}{0.018542} & \multirow{2}{*}{0.00102} & \multirow{2}{*}{0.001312} & \multirow{2}{*}{0.005501} & \multirow{2}{*}{0.027179}  \\ 
\ (sec) \ & & & & &  \\ \hline

\end{tabular}
}
\label{table:inference-speed-per-FusionConv}
\vspace{3mm}
\end{table*}
\begin{table*}[!t]
\centering
\caption{\textbf{Inference time of recent methods~\cite{stereolidar_00,stereolidar_norm_costV_ccvn} baseline networks~\cite{gcnet,psmnet}, and ours}}
\vspace{-2mm}
\resizebox{0.8\linewidth}{!}{
\begin{tabular}{|c|c|c|c|c|c|}
\hline
 & Park~\etal~\cite{stereolidar_00} & GC-Net~\cite{gcnet} & PSMNet~\cite{psmnet} & \ CCVN~\cite{stereolidar_norm_costV_ccvn} \ & \ \ \ Ours \ \ \  \\ \hline
Inference time (sec) & 0.043 & 0.962 & 0.985 & 1.011 & 1.40 \\ \hline
\end{tabular}}
\label{table:inference-time-recent-methods}
\vspace{3mm}
\end{table*}
\begin{figure*}[!t]
\centering
\includegraphics[width=1.0\linewidth]{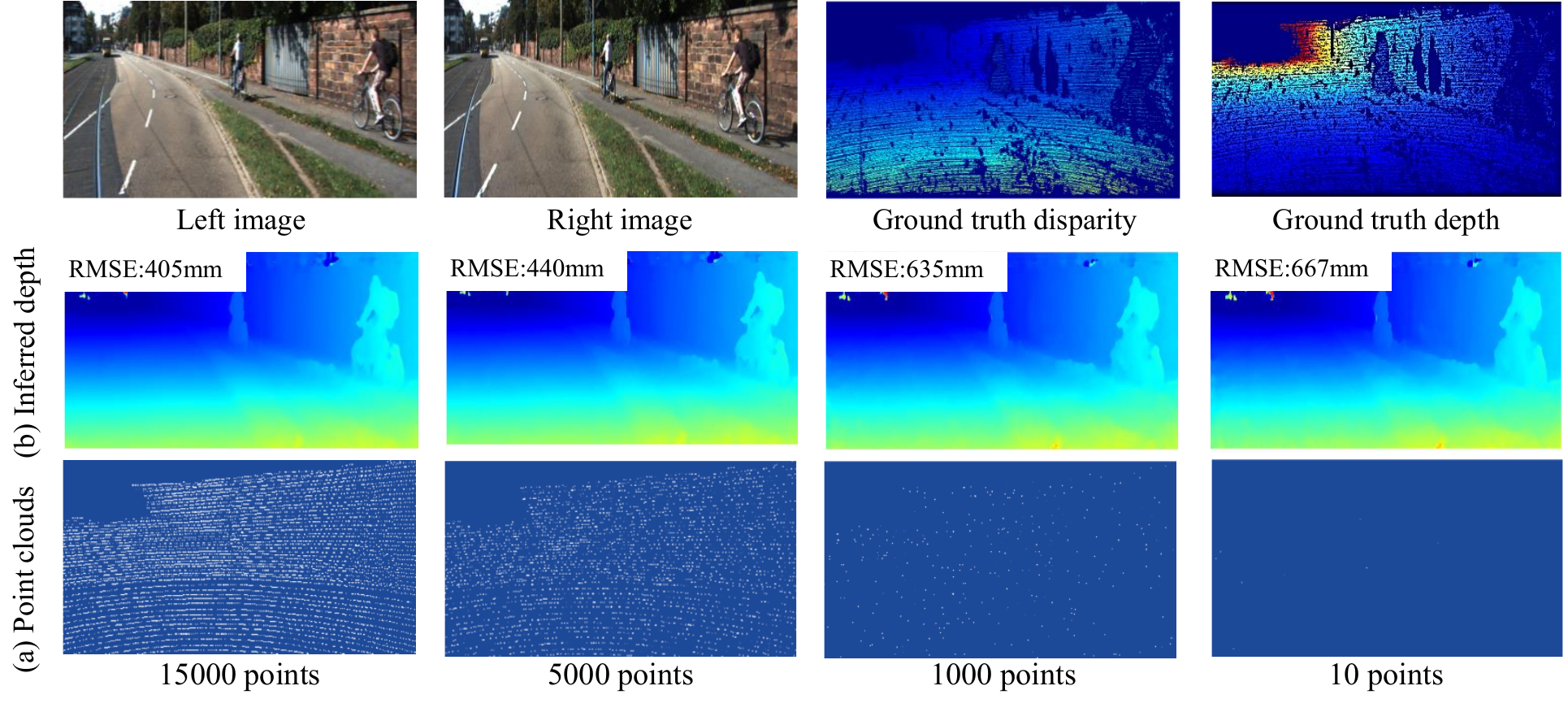}
\vspace{-3mm}
\caption{\textbf{Illustration of an ablation study for the quality of depth from different numbers of points.} We utilize one specific scene in the KITTI depth completion validation benchmark~\cite{kitti-completion} to measure the depth performance in 4 different cases of (a) input point clouds: 15000 points, 5000 points, 1000 points, and 10 points. In spite of reduced scale of input points, our method can estimate depth maps from stereo images functionally.}
\label{fig:fig_lidar}
\end{figure*}

\section{LiDAR dependency}
\label{supp-sec:LiDAR dependency}
In a real-world scenario, LiDAR provides not only sparse but variable points as ray fails at transparent, highly reflective regions. The stereo-LiDAR system can be one solution since it takes 3D information from both stereo images and point clouds. To validate the dependency of point clouds, we newly conduct an ablation study of the number of point clouds for stereo-LiDAR fusion as shown in~\Fref{fig:fig_lidar}. We randomly remove points and the remaining points are used for depth estimation. Using our trained network, we measure the depth performance in 4 different cases: 15000 points, 5000 points, 1000 points, and 10 points. Obviously, the fewer points we use, the less depth accuracy we have. However, with the visualized quality of inferred depth maps, it is reasonable to say that our method is not fully dependent on the point clouds, and is robust to the different number of point clouds for depth estimation.


\begin{figure*}[!t]
\centering
\includegraphics[width=0.9\linewidth]{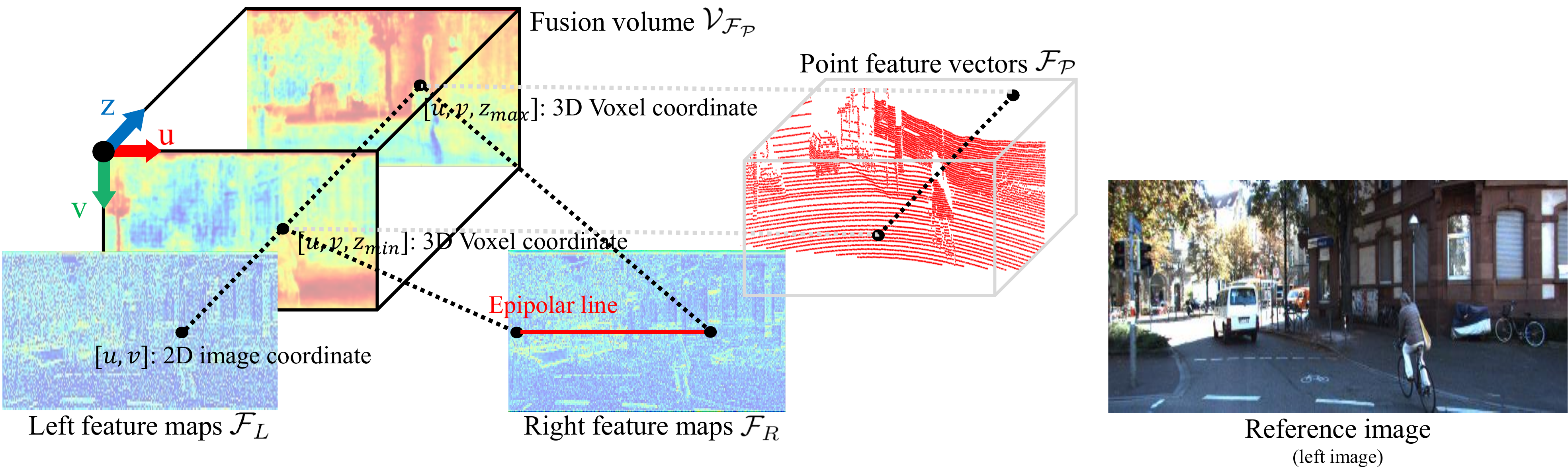}
\vspace{-2mm}
\caption{\textbf{Visualization of the composition of the fusion volume.} 
Under the known calibration parameters and the baseline between stereo cameras, there is a geometric relation between the location of the voxel $[u,v,z]$ and the feature data from stereo images and point clouds. This geometric fact is utilized for the composition of the fusion volume.}
\vspace{-2mm}
\label{fig:fig-volume}
\end{figure*}
\section{Fusion volume composition}
\label{supp-sec:Fusion volume composition}
This section describes the details of the composition of the fusion volume that we explain in~\Sref{sec:Volumetric Propagation} of the manuscript. Our fusion volume is the 3D volumetric data that represents the referential camera ray from $z_{min}$(${=}0m$) to $z_{max}$(${=}100m$) in a metric scale $[u,v,z]$ as shown in~\Fref{fig:fig-volume}. From the pre-defined location of each voxel $[u,v,z]$ in the fusion volume $\mathcal{V}_{\mathcal{F}_{\mathcal{P}}}$, we construct the fusion volume using stereo image features ($\mathcal{F}_{L}, \mathcal{F}_{R} \in \mathbb{R}^{C \times H \times W}$) and point feature vectors ($\mathcal{F}_{\mathcal{P}} \in \mathbb{R}^{C \times N}$). The pre-defined position of each voxel $[u,v,z]$ can be projected into the right feature maps by the known camera intrinsic matrix $\mathbf{K}$. In addition, conversion of the voxel coordinate $[u,v,z]$ into the lidar coordinate $[x,y,z]$ can be calculated through the given calibration parameters in the KITTI Completion dataset~\cite{kitti-completion}.

To do so, our network learns to aggregate the fusion volume to find the correspondence between stereo images and point clouds (\ie, triangulation). 

\section{FusionConv}
\label{supp-sec:FusionConv}
In this section, we describe the details of FusionConv by comparing it with the recent point-based architecture studies~\cite{continuous_conv,dynamic_graph_conv,interp_conv,pointnet}. Our method can be viewed as in-between the two recent point-based architecture methods, the parametric continuous convolution network (PCCN~\cite{continuous_conv}) and the interpolated convolution layer (InterpConv~\cite{interp_conv}). FusionConv is base on these two pioneering studies~\cite{continuous_conv,interp_conv}. However, the main difference is a \textbf{different sampling strategy} which is related to what we described in Sec.~\textcolor{red}{\RNum{1}} of the manuscripts, -- \textit{aligning different sensor data into a unified space is an essential step to fully operate depth estimated task}. For a clear answer, here we describe the detailed difference from recent point-based architecture methods~\cite{continuous_conv,interp_conv} with respect to algorithmic viewpoint.

The PCCN~\cite{continuous_conv} utilizes K-nearest neighbors (K-NN), which is a \textbf{fixed number of neighboring points}, for sampling neighboring point clouds. In a real-world scenario, most of the point clouds are located near a LiDAR sensor, and densely crowded points can have limited representations when K-NN based methods are utilized.

To overcome this problem, InterpConv~\cite{interp_conv} develops an idea of interpolation to apply convolutional operation into \textbf{dynamic numbers of point clouds in the 3D metric space}. This method pre-defines the size of the cube-shaped windows and all points within these windows are considered as neighboring points to be convolved. This window-based sampling concept can consider various numbers of point clouds, which the previous work~\cite{continuous_conv} colud not do. Despite of the improvement, InterpConv~\cite{interp_conv} has an issue in the fusion of images and point clouds. 

This cube-shaped window in InterpConv has different receptive fields in the homogeneous pixel coordinate. As we discussed in Sec.~\textcolor{red}{\RNum{1}} of the manuscript, aligning different sensor data into a unified space is an essential step to fully operate a depth estimation task from stereo images and point clouds. To align two modalities in a unified volume (fusion volume) whose coordinate obeys $[u,v,z]$, we need to re-design the windows to align point clouds ($[x,y,z]$) into the fusion volume ($[u,v,z]$). From the point of view of geometry, \textbf{FusionConv samples numerous points along the reference camera ray and its neighbor (\ie, frustum-shape windows). To do so, the fused point clouds can have receptive fields that are similar to those of image feature maps.}

The details of the FusionConv are close to those of the previous works~\cite{continuous_conv,interp_conv}. Our FusionConv, as its name implies, is designed for the fusion of point clouds and images in a unified volumetric space $[u,v,z]$, which has not yet been fully studied in point clouds based methods~\cite{continuous_conv,interp_conv,pointnet,dynamic_graph_conv} and stereo-LiDAR fusion methods~\cite{stereolidar_00,stereolidar_01,stereolidar_norm_costV_ccvn}. 

\section{Evaluation metrics}
\label{supp-sec:Evaluation metrics}
In Tables~\textcolor{blue}{\RNum{1}} and~\textcolor{blue}{\RNum{2}} of the manuscript and~\Tref{table:ablation-range} of the supplementary material, our method shows higher accuracy for AbsRel, SqRel, RMSE, and MAE, whereas the previous work, CCVN~\cite{stereolidar_norm_costV_ccvn}, shows reasonable inverse depth quality for iRMSE and iMAE. This fact suggests that our method can estimate accurate depth information at the distant area, but the CCVN~\cite{stereolidar_norm_costV_ccvn} shows strength for the closer area. This property comes from our design choice of the depth range in the fusion volume, which can affect the depth performance in close or farther areas. 
This aspect can be viewed as a deficiency of our method, but this result is consistent with our intention. Our volumetric propagation network is designed to maintain the metric accuracy in point clouds, especially points at a distant region where a pair of stereo images has difficulty in finding the correspondence. Inverse depth metrics (iRMSE and iMAE) and depth quality (AbsRel, SqRel, RMSE, and MAE) are currently a trade-off. We expect a future work to seamlessly describe the 3D scene and show the highest quality in both depth and inverse depth metrics.


\end{document}